\documentclass[10pt,twocolumn,letterpaper]{article}

\usepackage{icb}
\usepackage{times}
\usepackage{epsfig}
\usepackage{graphicx}
\usepackage{amsmath}
\usepackage{amssymb}
\usepackage{booktabs}
\usepackage{multirow}
\usepackage{subfigure}
\usepackage{verbatim}
\usepackage[norule,symbol,perpage]{footmisc} 

\usepackage[pagebackref=true,breaklinks=true,letterpaper=true,colorlinks,bookmarks=false]{hyperref}

\icbfinalcopy 

\ificbfinal\fi

\makeatletter
\def\ps@IEEEtitlepagestyle{
	\def\@oddfoot{\mycopyrightnotice}
	\def\@evenfoot{}
}
\def\mycopyrightnotice{
	{}
}
\makeatother

\begin{document}

\title{Face Recognition from Sequential Sparse 3D Data via Deep Registration}
\author{Yang Tan\textsuperscript{1,2}, Hongxin Lin\textsuperscript{1,2}, Zelin Xiao\textsuperscript{1,2}, Shengyong Ding\textsuperscript{*}, Hongyang Chao\textsuperscript{1}\\
	\textsuperscript{1}School of Data and Computer Science, Sun Yat-sen University\\
	\textsuperscript{2}Pixtalks Tech\\
	{\tt\small \{tany36,linhx9,xiaozl\}@mail2.sysu.edu.cn, marcding@163.com, isschhy@mail.sysu.edu.cn}
}	
\maketitle

\renewcommand{\thefootnote}{\fnsymbol{footnote}}
\footnotetext[1]{Corresponding author.}

\begin{abstract}
	Previous works have shown that face recognition with high accurate 3D data is more reliable and insensitive to pose and illumination variations. Recently, low-cost and portable 3D acquisition techniques like ToF(Time of Flight) and DoE based structured light systems enable us to access 3D data easily, e.g., via a mobile phone. However, such devices only provide sparse(limited speckles in structured light system) and noisy 3D data which can not support face recognition directly. In this paper, we aim at achieving high-performance face recognition for devices equipped with such modules which is very meaningful in practice as such devices will be very popular. We propose a framework to perform face recognition by fusing a sequence of low-quality 3D data. As 3D data are sparse and noisy which can not be well handled by conventional methods like the ICP algorithm, we design a PointNet-like Deep Registration Network(DRNet) which works with ordered 3D point coordinates while preserving the ability of mining local structures via  convolution. Meanwhile we develop a novel loss function to optimize our DRNet based on the quaternion expression which obviously outperforms other widely used functions. For face recognition, we design a deep convolutional network which takes the fused 3D depth-map as input based on AMSoftmax model. Experiments show that our DRNet can achieve rotation error $0.95^\circ$ and translation error $0.28mm$ for registration. The face recognition on fused data also achieves rank-1 accuracy $99.2\%$ , FAR-0.001 $97.5\%$ on Bosphorus dataset which is comparable with state-of-the-art high-quality data based recognition performance. 
\end{abstract}

\let\thefootnote\relax\footnotetext{\mycopyrightnotice} 

\section{Introduction}
\label{Introduction}

\begin{figure}[t]
	\centering 
	\includegraphics[width=0.9\linewidth]{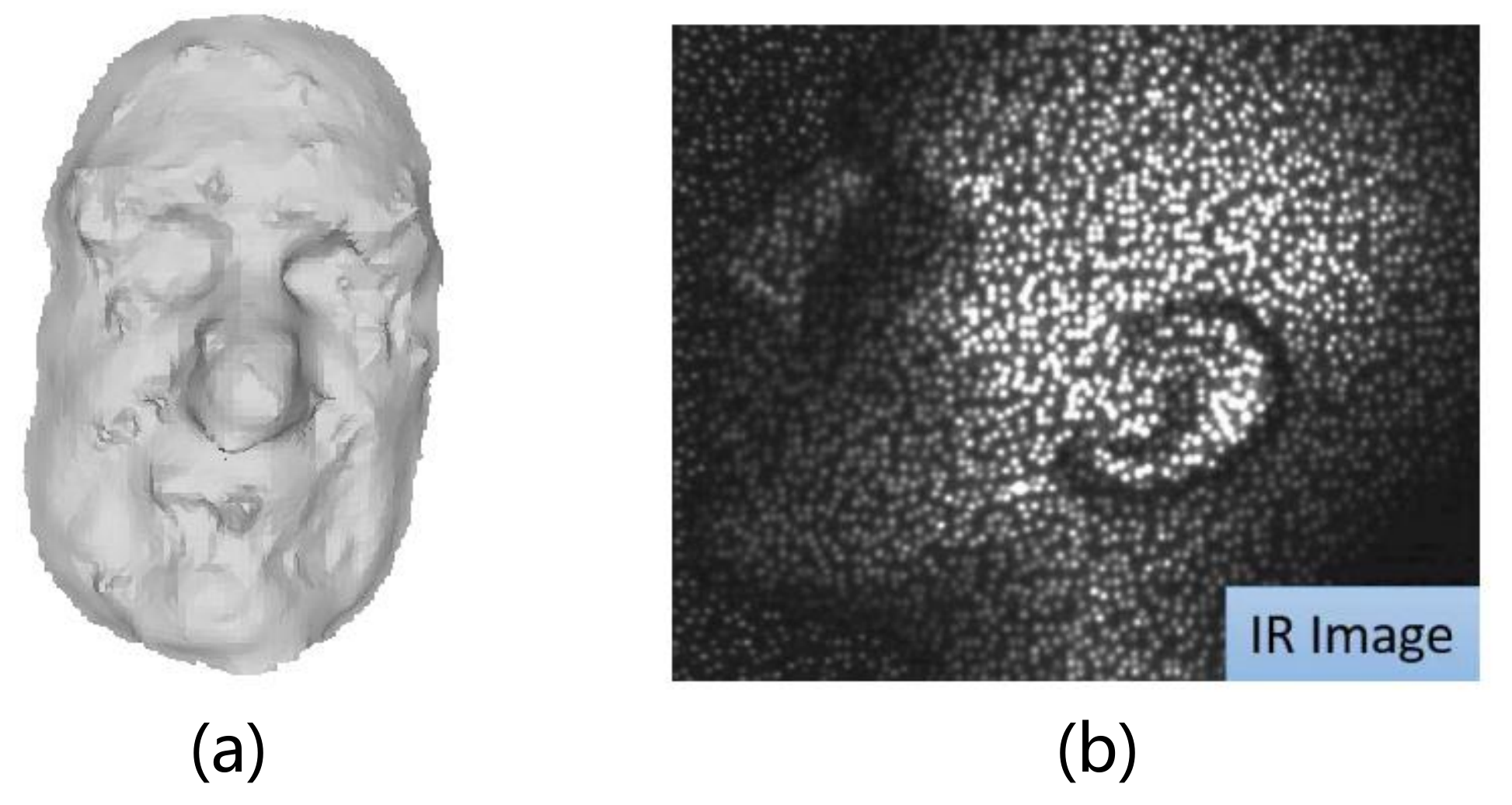} 
	\caption{(a) The reconstructed face model from low-quality 3D data (around 1,000 points). (b) The IR image acquired by DoE based structured light system.}
	\label{Fig1}  
	\end{figure}
	Face recognition from 3D data has been attractive due to its inherent advantages of being insensitive to pose and illumination variations. Previous literatures \cite{Gilani_2018_CVPR,Kim2017Deep} have shown that 3D face recognition from high-accuracy 3D data outperforms 2D recognition on a set of datasets. Recently, with the development of compact 3D acquisition techniques, e.g., time-of-flight(ToF) and DoE based structured light devices, people can easily access 3D data even with a mobile phone like iPhone X. The characteristic of such devices is that they can acquire 3D data at high frame speed while the quality of single frame is relatively poor, i.e., sparse and noisy. Figure \ref{Fig1}(a) shows the reconstructed face model from low-quality data. For instance, structured light systems derive the depth information by observing the speckle disparity between the reference pattern and the projected pattern on a surface. Due to the limit number of emitters, usually there are only hundreds of speckles projected onto a face. Figure \ref{Fig1}(b)  \cite{Fanello2016HyperDepth} shows DoE based patterns projected on a face. A more accurate and computation effective approach is to derive the depth from speckles but it leads to a sparse point cloud.\par
	In this paper, we aim at achieving high-performance face recognition for ToF and structured light based acquisition systems by using the fused sequential 3D data rather than a single frame. This is reasonable as the recognition process usually takes tens of milliseconds during which we can obtain several frames. Previous works \cite{Berretti2014Face,Lee2016Accurate} also tried to fuse frames acquired by Kinect to obtain a super-resolved model for face recognition. Note that the Kinect is not specifically designed for short distance scanning, so the face point cloud acquired by Kinect(dense but extremely noisy) is far from the new generation structured light devices(sparse but more accurate) designed for face recognition. In addition, they adopted common 3D registration algorithms like ICP \cite{Besl1992A} to align frames. The main shortcoming of such registration algorithms is incapable of handling large pose variations. Actually, in our experiments, we find that ICP algorithm almost can not handle difficult frame pairs when relative rotation angles $\alpha> 60^\circ,\gamma > 40^\circ$, where $\alpha,\gamma$ represent the roll and yaw angles.\par 
	Inspired by the success of Deep Learning in many vision tasks, we develop a PointNet-like \cite{Charles2017PointNet} Deep Registration Network(DRNet) to regress the registration parameters between any point cloud pair. More precisely, we hope our neural network takes a pair of interpolated point clouds as input and outputs a vector from which we can derive the transformation parameters. As translation parameters do not have any particular constraints, we simply regress translation parameters by $L_2$ loss. However, it is much more complex for rotation. The rotation can be expressed in several ways, e.g., a rotation matrix(nine parameters) $\mathbf{R}$, Euler angles$(\alpha, \beta, \gamma)$, axis-angle$(\theta,v_x,v_y,v_z)$ and quaternion$(\cos{\frac{\theta}{2}},\sin{\frac{\theta}{2}}\cdot v_x,\sin{\frac{\theta}{2}}\cdot v_y,\sin{\frac{\theta}{2}}\cdot v_z)$ where $(v_x,v_y,v_z)$ is the rotation axis and $\theta$ is the rotation angle. Considering the orthogonal constraints of rotation matrix and the  non-unique parametrisation of Euler angles \cite{Kendall2017Geometric}, we do not adopt these two expressions. Thus we design our loss function based on the unit quaternion system from the following facts. First, if a rotation is very small, then the rotation angle must also be very small, i.e. $\cos{\frac{\theta}{2}} \rightarrow 1$ no matter what the rotation axis is. Second, if two rotations $\mathbf{q}_1, \mathbf{q}_2$ are close, then the compositional rotation $\mathbf{q}_3=\mathbf{q}_1\mathbf{q}^{-1}_2$ must be small, indicating the real part of $\mathbf{q}_3$ approaching 1. Thus we define a loss function measuring the rotation angle between predicted and ground-truth pose for optimization. We find this loss function obviously outperforms other straight forward $L_2$ loss function on axis-angle$(\theta,v_x,v_y,v_z)$ expression.\par
	For face recognition, we design a convolutional neural network FRNet to achieve high recognition performance on our fused sequential data. Unlike the models used by \cite{Kim2017Deep,Gilani_2018_CVPR}, we adopt ResNet-18 \cite{He2016Deep} structure with AMSoftmax \cite{Wang2018Additive} loss function and find that simple augmentations for training data, e.g., pose variations and occlusions can surprisingly give good results without synthesizing any new identity. As there are no large scale existing sequential 3D datasets, we propose a method to generate desired data by sparse sampling and adding perturbations from existing high-quality face datasets. This approach enables us to obtain large amount of data at a very low cost while still producing meaningful results. \par 
	In summary, our contributions are follows:
	\begin{itemize}
		\item We raise a new challenging face recognition problem, i.e., face recognition from a sequence of sparse point clouds which will be common for structured light systems. Note the number of 3D points is only about 1,000 in our problem and much less than previous works \cite{Gilani_2018_CVPR,Kim2017Deep,Lee2016Accurate}.
		\item To handle large pose variations, we design a robust deep PointNet-like network DRNet for 3D point clouds rigid registration based on the unit quaternion expression with a carefully designed loss function. To the best of our knowledge, we are the first to align 3D facial point clouds via a neural network. 
		\item We study how the 3D data quality impacts face recognition and demonstrate the possibility of achieving high recognition accuracy from very sparse point cloud sequence.
		\end{itemize}
\section{Related work}
\label{Related work}
\begin{figure*}[tp]
	\centering 
	\includegraphics[width=1.0\linewidth]{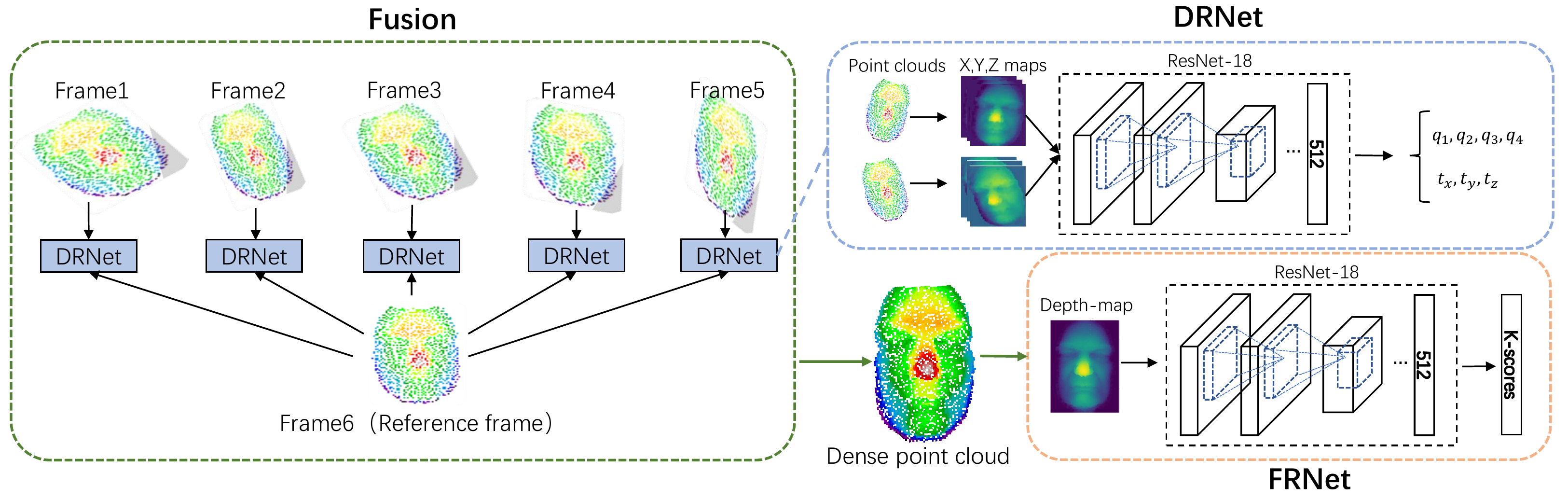} 
	\caption{An overview of our Face Registration and Face Recognition framework.} 
	\label{Fig2} 
\end{figure*} 
	 \textbf{Rigid Point Cloud Registration} The classical method for rigid point cloud registration is ICP \cite{Besl1992A} algorithm which aims at closing a pair of point clouds iteratively. However, the performance of ICP algorithm is heavily relied on the initial poses so it can not handle large pose variations. To address this problem, \cite{makadia2006fully} adopted EGI to perform coarse registration and then refined the result by ICP, but the complexity was unacceptable for mobile devices. \par
	 At present, there is no specifically designed neural networks for point cloud registration. FacePoseNet \cite{Chang2017FacePoseNet} directly regressed 6DoF transform parameters between the generic 3D facial keypoints model and the keypionts on intensity images. \cite{Kendall_2015_ICCV} proposed a network to estimate camera poses from monocular images in quaternion form. These methods are intensity image based but our method adopts coordinate-maps with accurate 3D informations as input and it seems more reasonable to predict 3D poses.\par
	 \textbf{3D Face Recognition} For conventional methods, there are local and global descriptor based approaches. Local descriptors usually extract some sub-regions and local informations as features. \cite{Li2015Towards} proposed a descriptor based on three face keypoints to describe local features. \cite{Gupta2010Anthropometric} performed 3D face recognition by comparing Euclidean and Geodesic distances between matched keypoints. Global descriptors treat face as an entity. \cite{Drira20133D} proposed using radial curves emanating from the nose tip to represent the facial surface. Some 3D Morphable Model \cite{Blanz1999A} based methods used 3DMM parameters for face recognition but the fitting process required massive computations.\par
	 CNN based methods DeepFace \cite{Taigman2014DeepFace} and FaceNet \cite{Schroff2015FaceNet} brought remarkable improvements for 2D face recognition. DeepFace achieved an accuracy of 97.35\% on LFW \cite{Fanello2016HyperDepth} dataset outperforming the best conventional method 27\% at that time. Later \cite{Kim2017Deep} proposed CNN based 3D face recognition pipeline and achieved comparable results. \cite{Gilani_2018_CVPR} proposed a synthesizing method to generate about 100 thousand identities for large scale training. These methods are trained on high-quality 3D data and can not be transfered to low-quality data directly.\par
	 \textbf{Sequential Methods} \cite{Yamaguchi1998Face} used a sequence of temporal images to perform 2D face recognition. Recent works \cite{Berretti2014Face,Lee2016Accurate,Hsu2014RGB,Choi2013Comparing} adopted a sequence of 3D data to perform depth fusion or morphology to reconstruct face models. They adopted ICP algorithm to perform registration for point clouds acquired by Kinect. As described in section \ref{Introduction}, Kinect data is far from our desired data due to the high noise level. We study the sparser but more accurate data acquired by the new generation 3D face scanner.   
	  
\section{Proposed framework}
\label{Proposed framework}
Our proposed framework is composed of two parts: Face Registration and Face Recognition. An overview of the framework is shown in Figure \ref{Fig2}. We first use Deep Registration Network(DRNet) to reconstruct a dense 3D facial point cloud from 6 frames of sparse point clouds by registrating and fusing. Then we feed the fused data to Face Recognition Network(FRNet).
\subsection{Deep Registration Network(DRNet)}
\label{Deep Registration Network(DRNet)}
\begin{figure}[t]
	\centering 
	\includegraphics[width=0.6\linewidth]{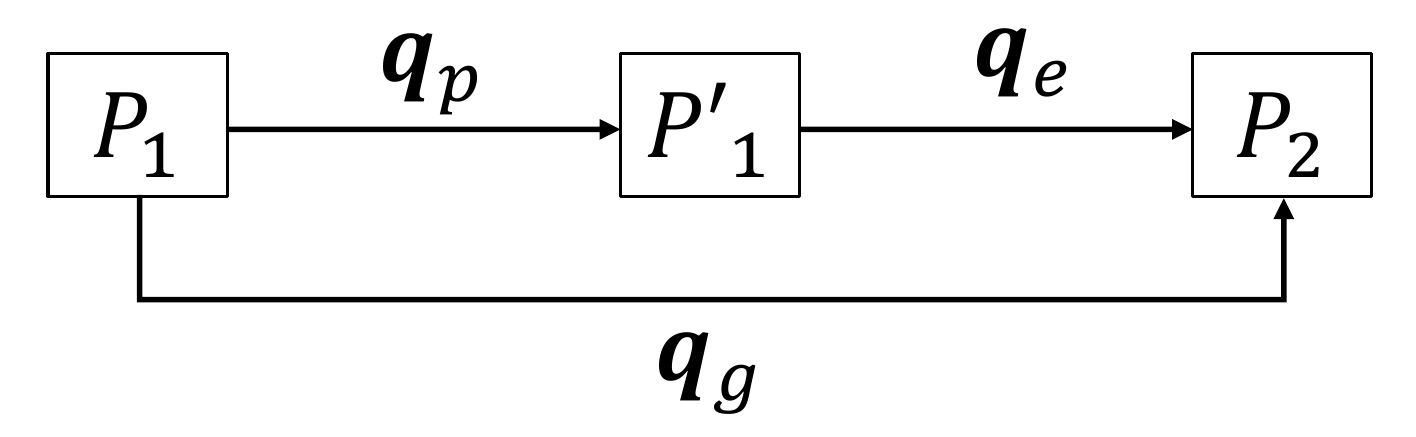} 
	\caption{$P_1,P'_1,P_2$ are source, predicted and target point clouds in different poses. $\mathbf{q}_g$ is the ground-truth rotation quaternion from $P_1$ to $P_2$. $\mathbf{q}_p$ is the predicted quaternion from $P_1$ to $P'_1$. $\mathbf{q}_e$ is the quaternion from $P'_1$ to $P_2$.} 
	\label{Fig3}  
\end{figure}
The most critical step to reconstruct the dense point cloud from a low-quality sequence is 3D registration, i.e., predict the rotation and translation transforms between two sparse point clouds. In order to handle large pose variations, we apply a deep neural network to regress the transformation parameters between two point clouds $P_1,P_2$. Inspired by PointNet \cite{Charles2017PointNet}, we introduce X,Y,Z coordinate-maps generated by projecting 3D points onto 2D image plane and then interpolating. Thus our network can be viewed as an implementation of 3D point cloud network which can still apply standard convolution and pooling operations to exploit hierarchical local structures. Our DRNet takes a pair of $256\times 256$ 3-channel coordinate-maps as input. The network architecture is based on ResNet-18 \cite{He2016Deep} with a 512-dimensional fully connected layer and outputs a 7-dimensional vector $\mathbf{p}$, encoding a translation vector $\mathbf{t} = (t_x,t_y,t_z)$ and a unit quaternion for rotation $\mathbf{q} = (q_1,q_2,q_3,q_4) \ s.t. \  q_1^2+q_2^2+q_3^2+q_4^2=1$:
\begin{equation}
\mathbf{p} = (\mathbf{t},\mathbf{q}) 
\end{equation}
\subsubsection{Interpretation of loss function}
\label{Interpretation of loss function}
The reason we express rotation in quaternion form is the nice properties of quaternion system, e.g., uniqueness and interpretability. The interpretability lies in the fact that the unit quaternion for a rotation $\mathbf q$ takes the following form:
\begin{equation}
\mathbf{q}= (\cos{\frac{\theta}{2}},\sin{\frac{\theta}{2}}\cdot v_x,\sin{\frac{\theta}{2}}\cdot v_y,\sin{\frac{\theta}{2}}\cdot v_z)
\end{equation}
with $(v_x,v_y,v_z)$ representing the rotation axis and $\theta$ being the rotation angle. Intuitively, we have two ways to construct the loss function, i.e., let the network directly regress quaternion$(\cos{\frac{\theta}{2}},\sin{\frac{\theta}{2}}\cdot v_x,\sin{\frac{\theta}{2}}\cdot v_y,\sin{\frac{\theta}{2}}\cdot v_z)$ or axis-angle $(\theta, v_x,v_y,v_z)$ by L1/L2 loss functions. In this paper, we design our loss from another aspect which is more interpretable and demonstrate that optimizing the quaternion form shows incomparable advantages. For ease of deduction, we first introduce some notations as shown in Figure \ref{Fig3} where $P_1$, $P'_1$ and $P_2$ represent the point clouds of the same rigid object in different poses.  We use the quaternion $\mathbf{q}_p$ to encode the rotation between point cloud $P_1$ and $P'_1$, $\mathbf{q}_g$ to encode the rotation between $P_1$ and $P_2$ and $\mathbf{q}_e$ to encode the rotation between $P'_1$ and $P_2$. In quaternion system, above definitions imply following equations:
\begin{equation}
\begin{aligned}
\mathbf{P}'_1 &= \mathbf{q}_p\mathbf{P}_1\mathbf{q}_p^{-1}  \\
\mathbf{P}_2 &= \mathbf{q}_g\mathbf{P}_1\mathbf{q}_g^{-1}  \\
\mathbf{P}_2 &= \mathbf{q}_e\mathbf{P}'_1\mathbf{q}_e^{-1}  \\
\Rightarrow  \mathbf{q}_e &= \mathbf{q}_g \mathbf{q}_p^{-1}
\end{aligned}
\end{equation}
According to the equation (2), it is not hard to see that if $\mathbf{q}_p$ exactly equals $\mathbf{q}_g$, then $\mathbf{q}_e = (1,0,0,0)$ implying there is no pose difference between $P'_1$ and $P_2$. So a well trained network should produce $\mathbf{q}_p$ with $\mathbf{q}_e = \mathbf{q}_g \mathbf{q}_p^{-1}$ approaching (1,0,0,0), thus the loss function can be defined as:
\begin{equation}
\begin{aligned}
loss_1 &= ||\textbf{q}_e - (1,0,0,0)||^2_2 \\
&=2 - 2\cos\frac{\theta_e}{2} 
\end{aligned}
\end{equation}
It demonstrates that the network has a clear optimizing goal in quaternion form, i.e., minimizing the rotation angle $\theta_e$ between predicted pose and target pose which has nothing to do with rotation axis. Actually, when the rotation angle approaches 0, we could say there is almost no rotation, no matter what the axis is. However, for another similar expression axis-angle$(\theta, v_x,v_y,v_z)$ also encoding the rotation angle and axis, the network has to optimize four variables simultaneously and it is difficult to judge which variable is more important.\par
Interestingly, the equation (4) $loss_1$ equals the common $L_2$ loss function on $\mathbf{q}_p$ defined as: 
\begin{equation}
\begin{aligned}
loss_2 &= ||\mathbf{q}_g - \mathbf{q}_p||^2_2\\
&= 2 - 2\cos{\frac{\theta_e}{2}}\\
&=loss_1
\end{aligned}
\end{equation}
For easy implementation, we define the complete loss function as:
\begin{equation}
loss = ||{\mathbf{t}_g} - \mathbf{t}_p||_2 + \alpha||\mathbf{q}_g - \mathbf{q}_p||_2
\end{equation} 
where $\alpha$ is a scale factor initialized as 500 to balance the translation and rotation weights. It increases to $1\times 10^{4}$ when $||\mathbf{q}_g - \mathbf{q}_p||^2_2 < 1\times10^{-4}$ during training. We adopt Adam \cite{Kingma2014Adam} optimizer with weight decay $5\times 10^{-5}$ and the initial learning rate is 0.01. Experiments clearly show that our loss function is obviously better than other intuitive loss functions such as $L_1$ loss on quaternion or axis-angle.
\subsection{Face Recognition Network (FRNet)} We aim at achieving high-accuracy face recognition with the fused sequential data which will be described in section \ref{Data generation}. We adopt the novel AMSoftmax \cite{Wang2018Additive} loss function to optimize our FRNet. The AMSoftmax loss function tries to separate different individuals on a sphere with a large margin. Our network takes the depth-map of size $256\times256$ converted as input. Again we use the ResNet-18 architecture with a 512 dimensional fully connected layer as the output feature. The facial similarity is measured by the simple cosine distance on output features. We train FRNet by Adam optimizer with weight decay $5\times 10^{-5}$ and the initial learning rate is 0.01. 

\section{Data generation}
\label{Data generation}
 As the new DoE based 3D scanners for face have not been mass produced, we are unable to construct a real large scale dataset to validate our method. Therefore, in this section, we will introduce the method to generate our simulated data from existing public 3D face datasets. To facilitate follow-up operations, we roughly align all face scans in datasets to a standard pose in advance.  
\subsection{Data for DRNet}
\label{Data for DRNet}
Face scans in raw datasets are dense(more than 10,000 points) and clean, so we need to perform sparse sampling to generate our desired data. To simulate the distribution of sparse and random patterns on a moving face, we develop a sparse sampling strategy. First, we introduce noises and a random pose variation which is expressed in Euler angles $\alpha \in [-45^\circ,45^\circ],\beta \in [-20^\circ,20^\circ],\gamma \in [-30^\circ,30^\circ]$ representing roll, pitch and yaw angles respectively and the translation $t_x,t_y,t_z \in [-8mm,8mm]$ to a pre-aligned face scan. Noises come from a Gaussian distribution $N(0,4)$ and are randomly added to ten percent of points. Second, we project the point cloud onto a 2D plane divided into 1,000 grids of the same size and randomly select one point from each grid, thus we obtain a sparse point cloud containing about 1,000 points. For each face scan in raw datasets, we repeat the procedure above 6 times to obtain a sequence of sparse face data, while one of six frames is defined as the reference frame with  $\alpha = 0^\circ,\beta =0^\circ,\gamma = 0^\circ$. Note that the pose variation is relative to the standard pose and transform parameters will be recorded to calculate the relative transform between any pair of frames in the sequence. \par
During the training stage, we randomly feed pairs of frames from the training sequential dataset to regress rotation and translation parameters.     
\subsection{Data for FRNet} 
\label{Data for FRNet}
Our goal is to demonstrate that the fused data from a sparse sequence of 6 frames can also achieve comparable results to the high-quality data. To study the performances of our FRNet under different data qualities, we generate 4 types of data as follows:
\begin{itemize}
	\item \textbf{Fused data} For each sparse sequence, we align other 5 frames to the reference frame by our DRNet, then the union of these 6 aligned point clouds is defined as fused data. As the fused data contains around 6,000 points, it is possible to conduct denoising. Specifically, for each 3D point $(x,y,z)$, we calculate the mean z-coordinate $z_m$ of neighbor points within $radius=3mm$. When $|z-z_m|>2$, update $z=z_m$. This denoising strategy can effectively remove outliers but does not smooth the point cloud excessively. 
	During the training stage, we first augment the fused data with pose variations $\alpha,\beta,\gamma \in[-10^\circ,10^\circ]$. Then we project point clouds onto image plane and interpolate it to generate depth-maps. We randomly occlude depth-maps with 1-6 patches of size in $[0,20]$ for augmentation.
	\item \textbf{High-quality data} We select raw point clouds(more than 10,000 accurate points) from 3D face datasets. We also adopt the same augmentation strategy described above to generate depth-maps.
	\item \textbf{Low-quality data} We select reference frames(around 1,000 points) from sparse sequences. We also adopt the same augmentation strategy described above to generate depth-maps.
	\item \textbf{Sequential data} We directly feed sparse sequences to FRNet without registration and fusion. We also adopt the same augmentation strategy described above for each frame to generate depth-maps. Note that we need to expand the input channels of FRNet from one to six.   
\end{itemize} 
\section{Experiments}
In this section, we first evaluate the performance of face registration by DRNet and then evaluate face recognition performances on different types of face data.
\subsection{Datasets}
Table \ref{tab1} shows the most popular 3D face datasets including ND-2006 \cite{Faltemier2007Using}, Bosphorus \cite{Savran2011Comparative}, CASIA \cite{Xu2006Learning} and UMBDB \cite{Colombo2011UMB}. The complete ND-2006 dataset is used to construct our training set. We select 2,900 scans(gallery 105, probes 2,795) from Bosphorus, 3,555 scans(gallery 123, probes 3,432) from CASIA, 749 scans(gallery 122, probes 627) from UMBDB except side faces and extremely occluded samples to construct testing sets.\par
Specially, we construct two testing sets for face registration, i.e., standard set and difficult set. Both of them contain 4,000 pairs of point clouds generated from Bosphorus dataset. The difference is that samples in standard set uniformly pose in $\alpha \in[-45^\circ,45^\circ],\beta \in[-20^\circ,20^\circ],\gamma \in[-30^\circ,30^\circ]$, but samples in difficult set pose in $\alpha \in[-45^\circ,-30^\circ]\cup[30^\circ,45^\circ],\beta \in[-20^\circ,20^\circ],\gamma \in[-30^\circ,-20^\circ]\cup[20^\circ,30^\circ]$, where  $\alpha,\beta,\gamma$ are Euler angles described in section \ref{Data for DRNet}
\begin{table}[h]
	\caption{Details of datasets}
	\renewcommand \tabcolsep{3.0pt}
	
	\begin{center}
		\begin{tabular}{|l|ccccc|}
			\hline
			Name & IDs & Scans & Expressions & Pose & Occlusion \\ 
			\hline\hline
			ND-2006 & 888 & 13,450 & Multiple & $\pm 15^\circ$ & None \\
			Bosphorus & 105 & 4,666 & 7 & $\pm 90^\circ$ & 4 types \\
			CASIA & 123 & 4,674 & 6 & Frontal & None \\
			UMBDB & 143 & 1,473 & 4 & Frontal & 7 types \\
			\hline
		\end{tabular}
	\end{center}
	\label{tab1}
\end{table}
 \subsection{Evaluation of Face Registration}
 We quantitatively evaluate the registration result by rotation error and translation error. The  rotation error  $\theta_e$ is defined as how many degrees still need to be rotated from the predicted pose to the ground-truth pose, which is derived from the real part of $\mathbf{q}_e = \mathbf{q}_g\mathbf{q}^{-1}_p$ shown in Figure \ref{Fig3}. The translation error is measured by $t_e = ||\mathbf{t}_g - \mathbf{t}_p||_2$ where $\mathbf{t}_g$ and $\mathbf{t}_p$ are the ground-truth translation and predicted translation. \par
\begin{figure}[htbp]
 	\centering 
 	\includegraphics[width=1.0\linewidth]{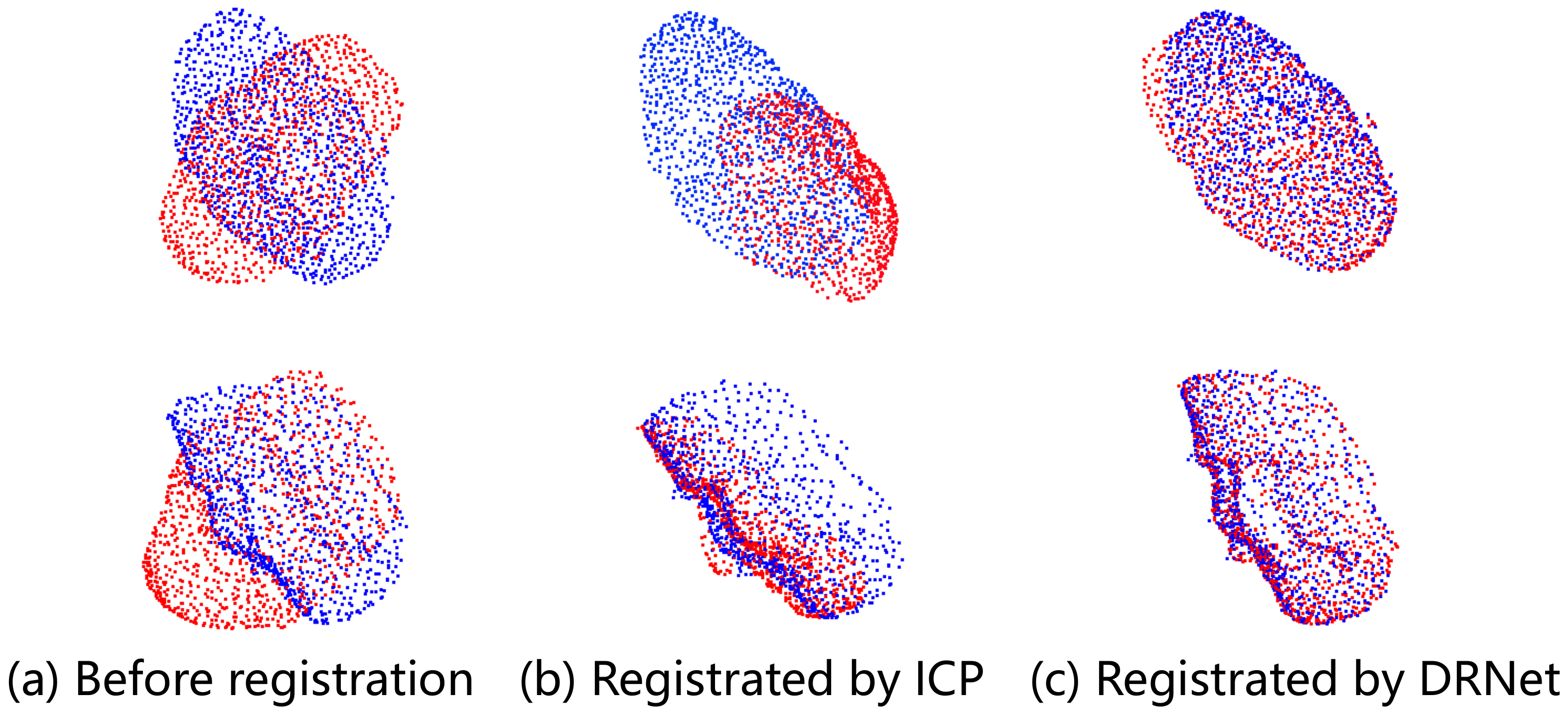} 
 	\caption{Two failure samples of ICP algorithm. (a) column shows two pairs of sparse point clouds before registration. Blue represents the target point cloud. (b) column shows the registration result of ICP with significant errors. (c) column shows the well aligned result by our DRNet,.} 
 	\label{Fig4}  
 \end{figure}
\begin{figure}[t]
	\centering 
	\includegraphics[width=1.0\linewidth]{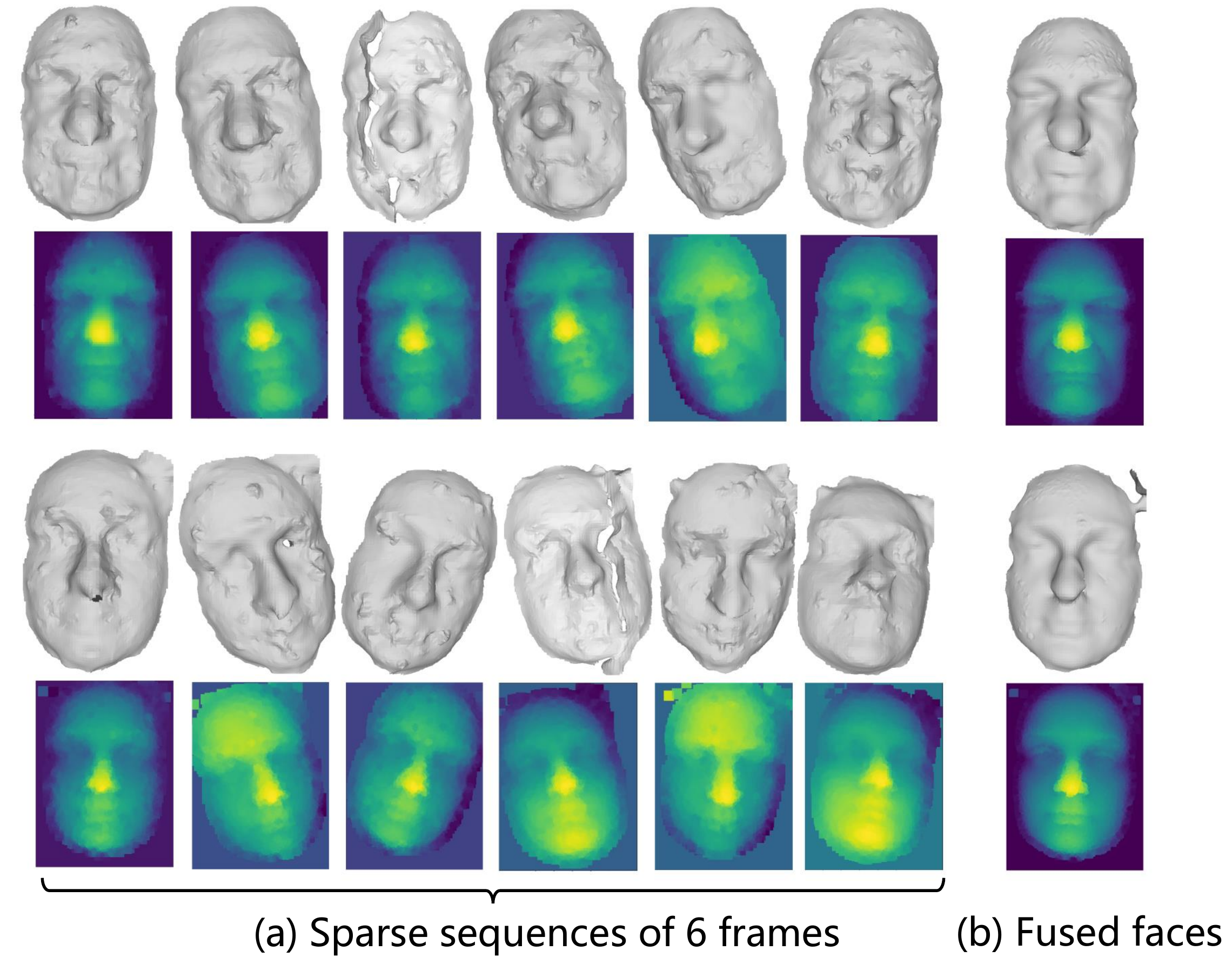} 
	\caption{(a) shows two sequences of low-quality 3D data. (b) shows the reconstructed faces after registration and fusion, which contain more facial details.} 
	\label{Fig5}  
\end{figure}

 Table \ref{tab2} shows that our DRNet achieves an impressive registration performance, especially on difficult testing set. We can see that both the ICP algorithm and our DRNet work well on standard testing set, i.e., the relative rotation angle between two point clouds satisfies  $\Delta\alpha<60^\circ$ and $
 \Delta\gamma<40^\circ$. However, for large pose variations, ICP algorithm usually falls into a local optimum and produces large errors. Actually, on difficult testing set, we find that about $11.8\%$ registrations fail with ICP algorithm, i.e., meaningless alignment as shown in Figure \ref{Fig4}, while none registration fails with DRNet. Note that on standard testing set, ICP algorithm performs slightly better than DRNet since ICP can recursively refine the registration results. We observe the same effect with DRNet, i.e., a second registration by DRNet will produce better results as shown in Table \ref{tab2}. In addition, Figure \ref{Fig5} shows fused faces from sequences of sparse data.\par
 As the discussion in section \ref{Interpretation of loss function}, a rotation can be expressed in axis-angle form $(\theta, v_x,v_y,v_z)$ and quaternion form $(\cos{\frac{\theta}{2}},\sin{\frac{\theta}{2}}\cdot v_x,\sin{\frac{\theta}{2}}\cdot v_y,\sin{\frac{\theta}{2}}\cdot v_z)$. We give the registration results on DRNets based on these two forms with L1 and L2 loss functions in Table \ref{tab3}. Experiments clearly show that Quaternion-L2 is superior to others as expected.
 \begin{table}[h]
 	\caption{Average registration errors on standard testing set and difficult testing set. }
 	\vspace{6pt}
 	\renewcommand \tabcolsep{10.0pt}
 	\begin{tabular}{|c|cc|cc|}
 		
 		\hline
 		\multirow{2}{*}{Method} &\multicolumn{2}{c}{Standard}&\multicolumn{2}{|c|}{Difficult}\cr
 		\cline{2-5}
 		&$\theta_e(^\circ)$ & $t_e (mm)$ & $\theta_e(^\circ)$ & $t_e (mm)$ \cr
 		\hline\hline
 		$ICP$    & \textbf{0.77}&0.27          & 7.64         &1.92 \cr
 		$Ours$   & 1.80         & 0.38         & 1.83         & 0.45  \cr
 		$Ours^*$ & 1.08         & \textbf{0.25}& \textbf{0.95} & \textbf{0.28}  \cr
 		\hline
 	\end{tabular}
 	\footnotesize{$^*$represents performing registration twice.}
 	\label{tab2}
 \end{table} 
 \begin{table}[h]
 	\centering
 	\caption{Average registration errors of different loss functions and expressions on standard testing set. }
 	\vspace{6pt}
 	\renewcommand \tabcolsep{18.0pt}
 	\begin{tabular}{|c|cc|}
 		\hline
 		Method &$\theta_e(^\circ)$ & $t_e (mm)$ \cr
 		\hline\hline
 		Axis-angle-L1  & 4.67 & 0.71  \cr
 		Axis-angle-L2  & 3.17 & 0.41  \cr
 		Quaternion-L1  & 2.79 & 0.60  \cr
 		Quaternion-L2  & \textbf{1.80} & \textbf{0.38}  \cr
 		\hline
 	\end{tabular}
 	\label{tab3}
 \end{table}
\subsection{Evaluation of Face Recognition}
\subsubsection{Settings} 
For comparison, we evaluate the performances of face recognition with data described in section \ref{Data for FRNet}. Below are experimental settings:
\begin{itemize}
\item$FRNet_{F}$: Use the fused data for training and testing. It is designed to demonstrate whether our fusion strategy is effective.
\item$FRNet_{H}$: Use the high-quality data for training and testing. It is designed to compare with the state-of-the-art high-quality data based methods and demonstrate whether our FRNet is well designed.
\item$FRNet_{H}^f$: Use the high-quality data for training but fused data for testing to study whether the high-quality data based model is capable of adapting to our fused data.
\item$FRNet_{L}$: Use the low-quality data for training and testing. We want to know if it is possible to achieve high performance recognition just from one single frame of sparse face data.
\item$FRNet_{S}$: Use the sequential data for training and testing. It is designed to demonstrate whether the registration is necessary.
\end{itemize}

\subsubsection{Performances} 
\begin{figure*}[t]

\subfigure[CMC curve on Bosphorus]{
	\includegraphics[width=0.33\textwidth]{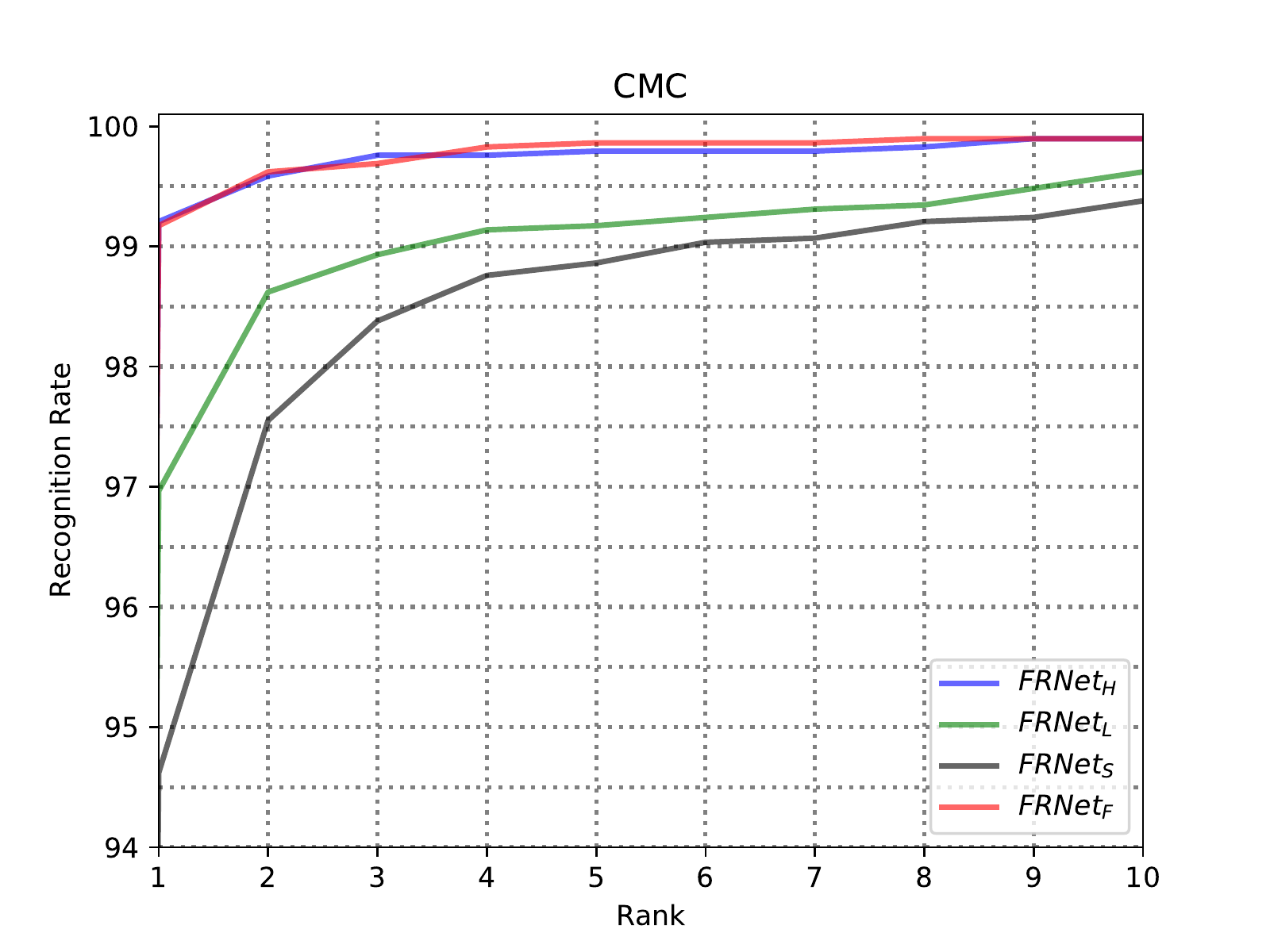}}
\subfigure[CMC curve on CASIA]{\includegraphics[width=0.33\textwidth]{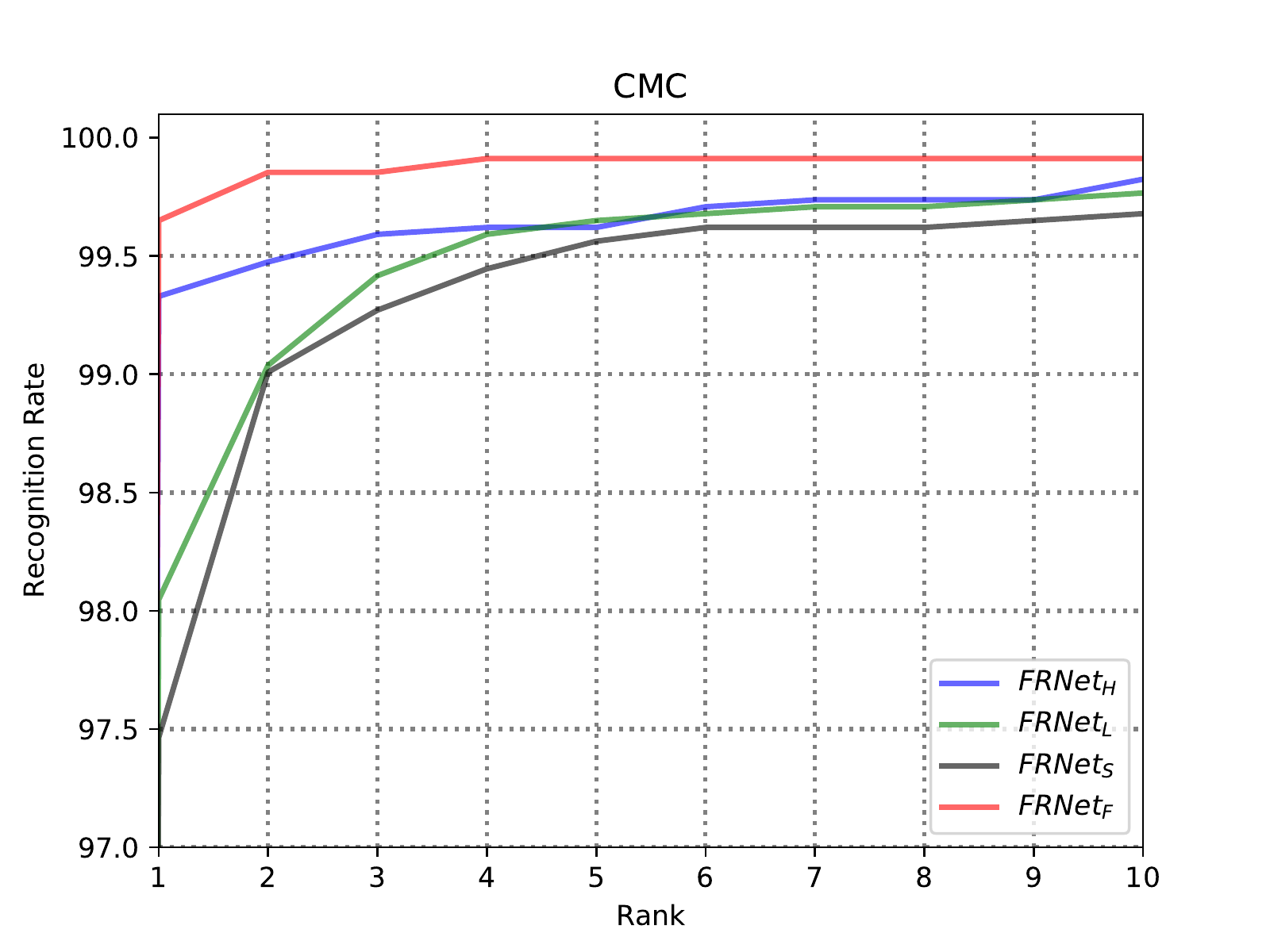}}
\subfigure[CMC curve on UMBDB]{\includegraphics[width=0.33\textwidth]{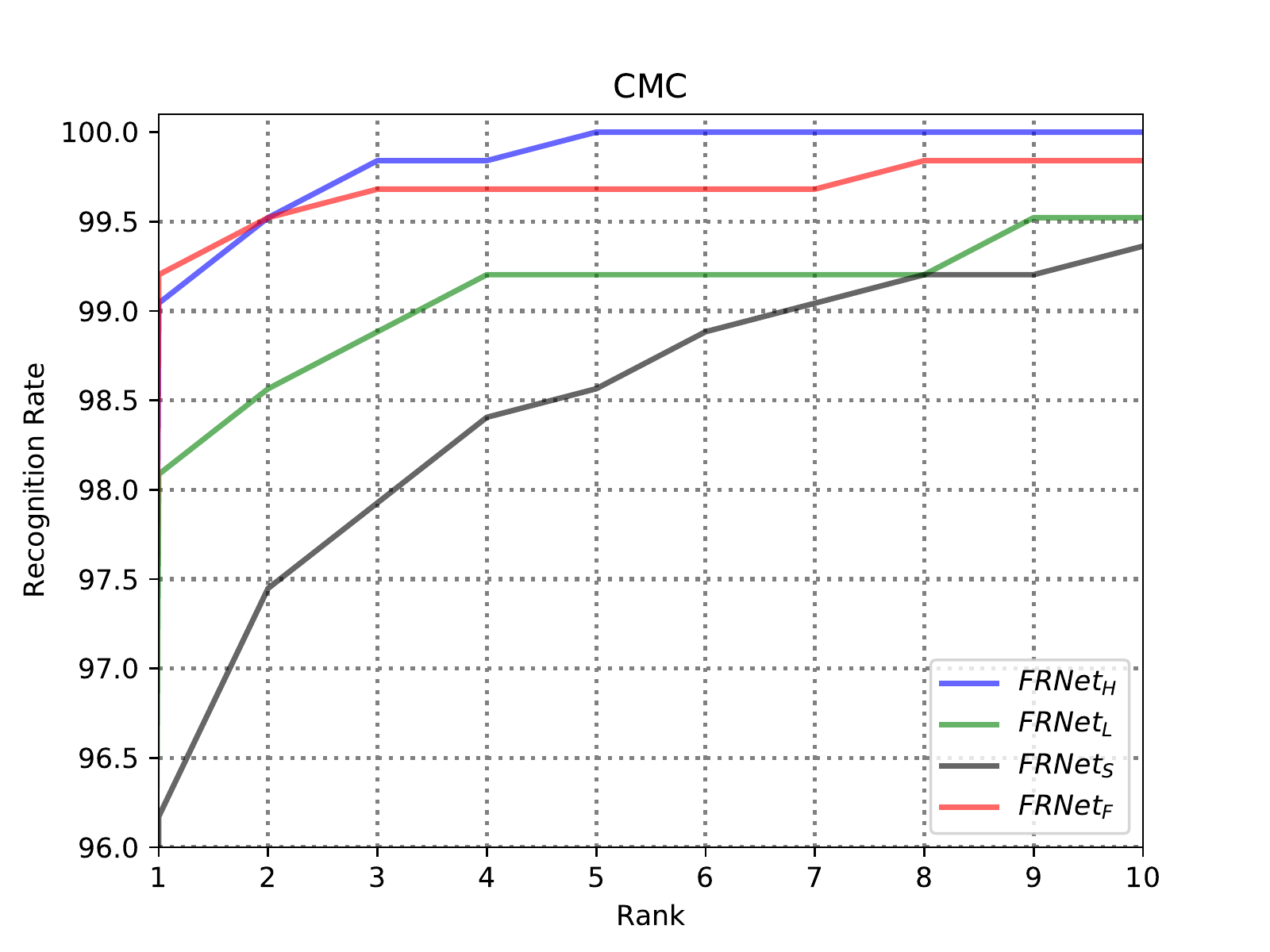}}
\vfill
\subfigure[ROC curve on Bosphorus]{\includegraphics[width=0.33\textwidth]{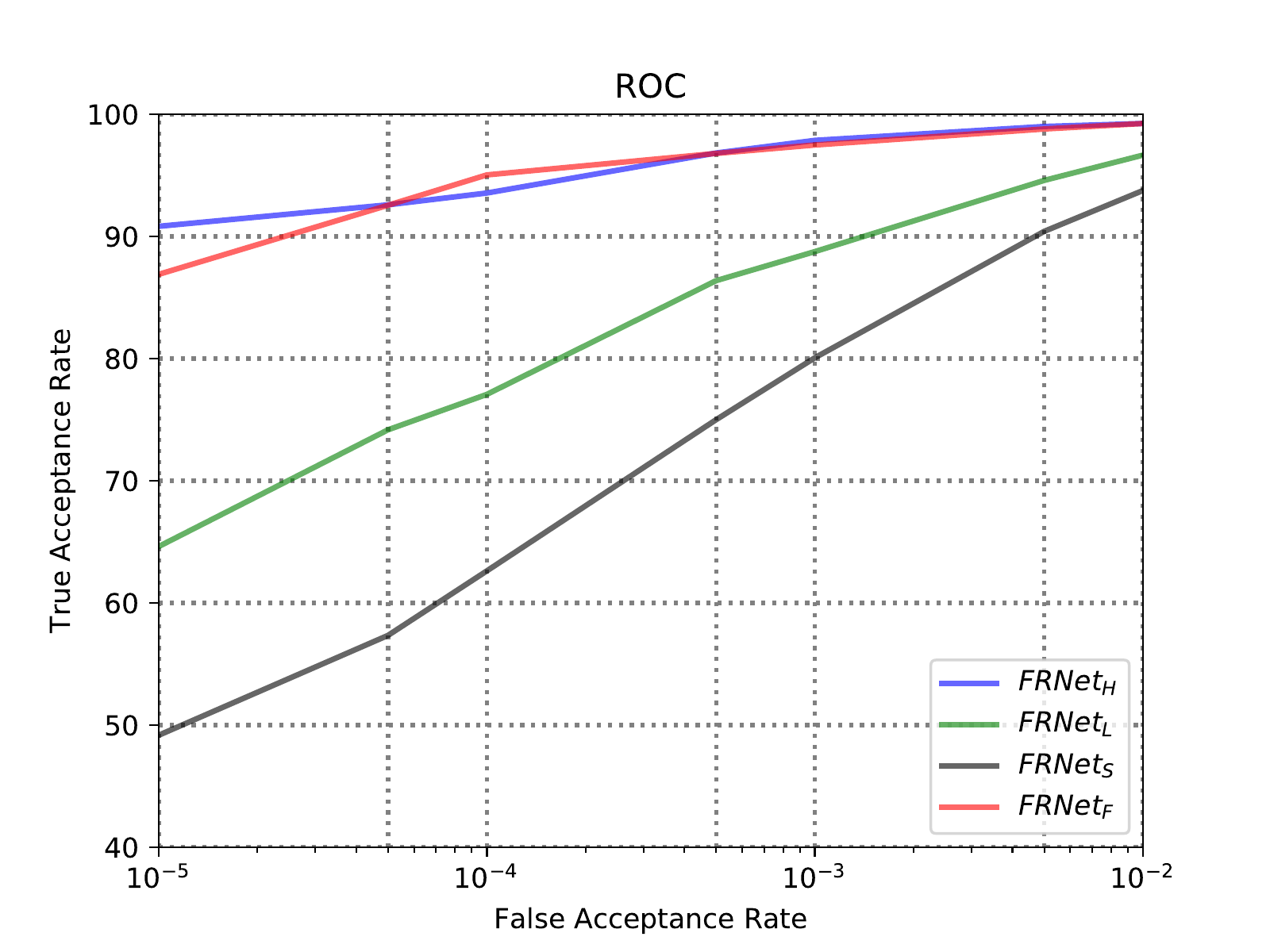}}
\subfigure[ROC curve on CASIA]{\includegraphics[width=0.33\textwidth]{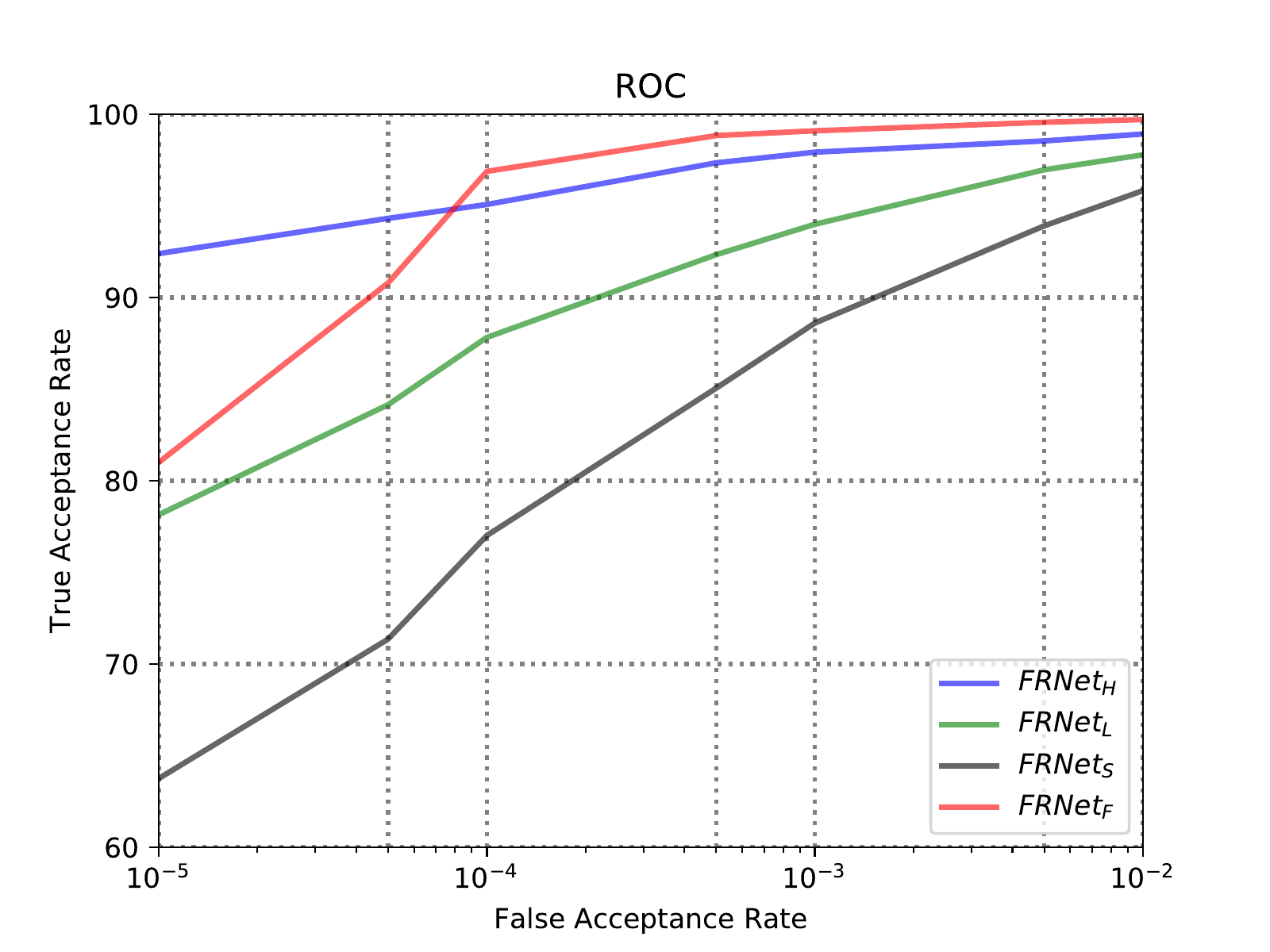}}
\subfigure[ROC curve on UMBDB]{\includegraphics[width=0.33\textwidth]{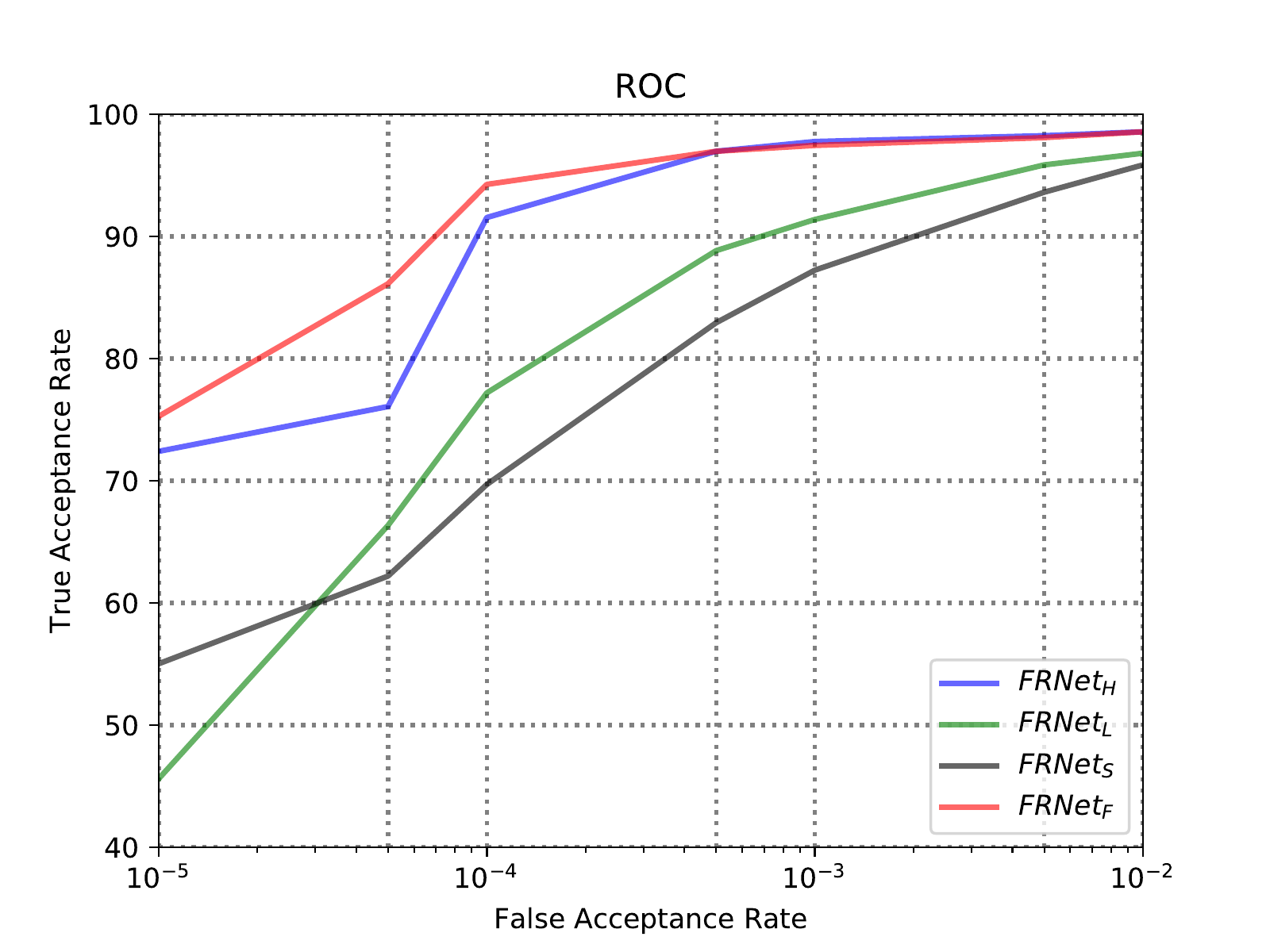}}
\caption{Face recognition performances on 3 testing sets}
\label {Fig6}
\end{figure*}

We follow the evaluation protocol described in \cite{Kim2017Deep} and adopt common criteria CMC and ROC. Figure \ref{Fig6} shows CMC and ROC curves on Bosphorus, CASIA and UMBDB testing sets. Table \ref{tab4} shows the result of False Acceptance Rate at 0.001. Table \ref{tab5} shows the rank-1 result compared with state-of-the-art methods. Firstly, our $FRNet_{H}$ and $FRNet_{F}$ achieve comparable and state-of-the-art results on testing sets, which shows that our fused data is capable of performing high-accuracy recognition as high-quality data. Note that the $FRNet_{F}$ is a little bit better than $FRNet_{H}$ on some criteria and we consider it may be caused by the noise of fused data which ease the over-fitting. Secondly, $FRNet_{L}$ just achieves Rank-1 $97.0\%$ and FAR-0.001 $88.8\%$ on Bosphorus which demonstrates that one single frame of sparse data is unable to achieve high recognition performance. Thirdly, the performances of $FRNet_{S}$ are also not really good, only achieving Rank-1 $94.6\%$ and FAR-0.001 $80.1\%$ on Bosphorus. We consider that it is hard for the network to utilize the complementary information from sequential frames without registration and fusion.
\begin{table}[h]
	\caption{Comparison of False Acceptance Rate (\%) at 0.001 with different settings}
	\vspace{6pt}
	\renewcommand \tabcolsep{11.0pt}
	\begin{tabular}[width=\linewidth]{|c|ccc|}
		\hline
		Method & Bosphorus & CASIA & UMBDB\\
		\hline\hline
		$FRNet_F$ & 97.5 & \textbf{99.1} & 97.4 \cr
		$FRNet_H$ & \textbf{97.9} & 97.9 & \textbf{97.8} \cr
		$FRNet^f_H$ & 96.8& 96.9 & 97.6 \cr
		$FRNet_L$ & 88.8 & 94.0 & 91.4 \cr
		$FRNet_S$ & 80.1 & 88.6 & 87.2 \cr
		\hline
	\end{tabular}
	\label{tab4}
\end{table}
\begin{table}[h]
	\centering
	\caption{Comparison of Rank-1 recognition accuracy (\%) with state-of-the-art methods}
	\vspace{6pt}
	\small
	\begin{tabular}[width=\linewidth]{|c|ccc|}
		\hline
		Method & Bosphorus & CASIA & UMBDB \cr
		\hline\hline
		Xu \etal~\cite{Xu2006Learning}(2006) & - & 83.9& - \cr
		Mian \etal~\cite{Mian2007An}(2007) & 96.4 & 82.5 & 69.3 \cr
		Gilani \etal~\cite{Zulqarnain2014Dense}(2018) & 98.6 & 85.4 & 78.6 \cr
		Lei \etal~\cite{Lei2016A}(2016) & 98.9 & - & - \cr
		Kim \etal~\cite{Kim2017Deep}(2017) & 99.2& - &- \cr
		Zulqarnain \etal~\cite{Gilani_2018_CVPR}(2018) & \textbf{100.0} & \textbf{99.7} & 97.2 \cr
		
		\hline
		$FRNet_F$ & 99.2 & \textbf{99.7} & \textbf{99.2} \cr
		$FRNet_H$ & 99.2 & 99.3 & 99.0 \cr
		$FRNet^f_H$ & 99.3 & 98.9 & 99.0 \cr
		$FRNet_L$ & 97.0 & 98.0 & 98.1 \cr
		$FRNet_S$ & 94.6 & 97.5 & 96.2 \cr
		
		\hline
	\end{tabular}
	\label{tab5}
\end{table}

\section{Conclusion}
In this paper, we propose a framework to achieve high-accuracy face recognition from sequential sparse and noisy 3D data. Unlike previous works relying on ICP algorithm for registration, we propose a deep convolutional network DRNet to regress the transformation parameters with a carefully designed loss function. Our DRNet is able to achieve rotation error $0.95^\circ$ and translation error $0.28mm$ even for large pose variations(difficult testing set). Using the fused data by DRNet for face recognition, we achieve rank-1 $99.2\%,99.7\%,99.2\%$ and FAR-0.001 $97.5\%,99.1\%,97.4\%$ on Bosphorus, CASIA and UMBDB datasets which is a comparable result with the performance tested on high-quality data.

{\small
\bibliographystyle{ieee}
\bibliography{FaceSeq_ICB}
}

\end{document}